\documentclass[10pt,twocolumn,letterpaper]{article}

\usepackage[pagenumbers]{cvpr} % To force page numbers, e.g. for an arXiv version
\usepackage{multirow}
\usepackage{bbding}
\usepackage{makecell}
\usepackage{indentfirst}

\definecolor{cvprblue}{rgb}{0.21,0.49,0.74}
\usepackage[pagebackref,breaklinks,colorlinks,allcolors=cvprblue]{hyperref}

\def\paperID{5401} % *** Enter the Paper ID here
\def\confName{CVPR}
\def\confYear{2025}

\title{Touch2Shape: Touch-Conditioned 3D Diffusion for \\ Shape Exploration and Reconstruction}

\author{
Yuanbo Wang$^1$ \hspace{0.5cm}
Zhaoxuan Zhang$^2$ \hspace{0.5cm}
Jiajin Qiu$^1$ \hspace{0.5cm}
Dilong Sun$^1$\\ 
Zhengyu Meng$^1$ \hspace{0.5cm}
Xiaopeng Wei$^1$\footnotemark[1] \hspace{0.5cm}
Xin Yang$^1$\footnotemark[1] \hspace{0.5cm}
\\
$^1$ Key Laboratory of Social Computing and Cognitive Intelligence, Dalian University of Technology \\
$^2$ Nanjing University of Posts and Telecommunications\\
\tt\small wangyuanbo@mail.dlut.edu.cn \hspace{0.5cm}
zhangzx@njupt.edu.cn\\
\tt\small \{jiajinqiu, sundilong, mzy1\}@mail.dlut.edu.cn \hspace{0.5cm}
\tt\small \{xpwei, xinyang\}@dlut.edu.cn
}

\begin{document}
% \maketitle
% %------------------------------------------------------------------------
% \begin{teaserfigure*}
%   \centering
%   \includegraphics[width=0.999\linewidth]{fig/teaser.png}
%   \caption{\label{fig:fig1}
%           From left to right: the architecture overview of our proposed models, event-intensity voxel fusion (EIVF) module, and event-intensity spatial pyramid fusion (EI-SPF) module. 
%           }
% % \vspace{-0.4cm} 
% \end{teaserfigure*}
% %----------------------------------------------------------------------------------

% \twocolumn[{%
% \maketitle
% \begin{figure}[H]
% \hsize=\textwidth % cvpr 需要
% \centering
% \includegraphics[width=3cm]{fig/teaser.png}
% \caption{foo}
% \end{figure}
% }]

\twocolumn[{
\renewcommand\twocolumn[1][]{#1}
\maketitle
\begin{center}
    \captionsetup{type=figure}
    \includegraphics[width=0.999\linewidth]{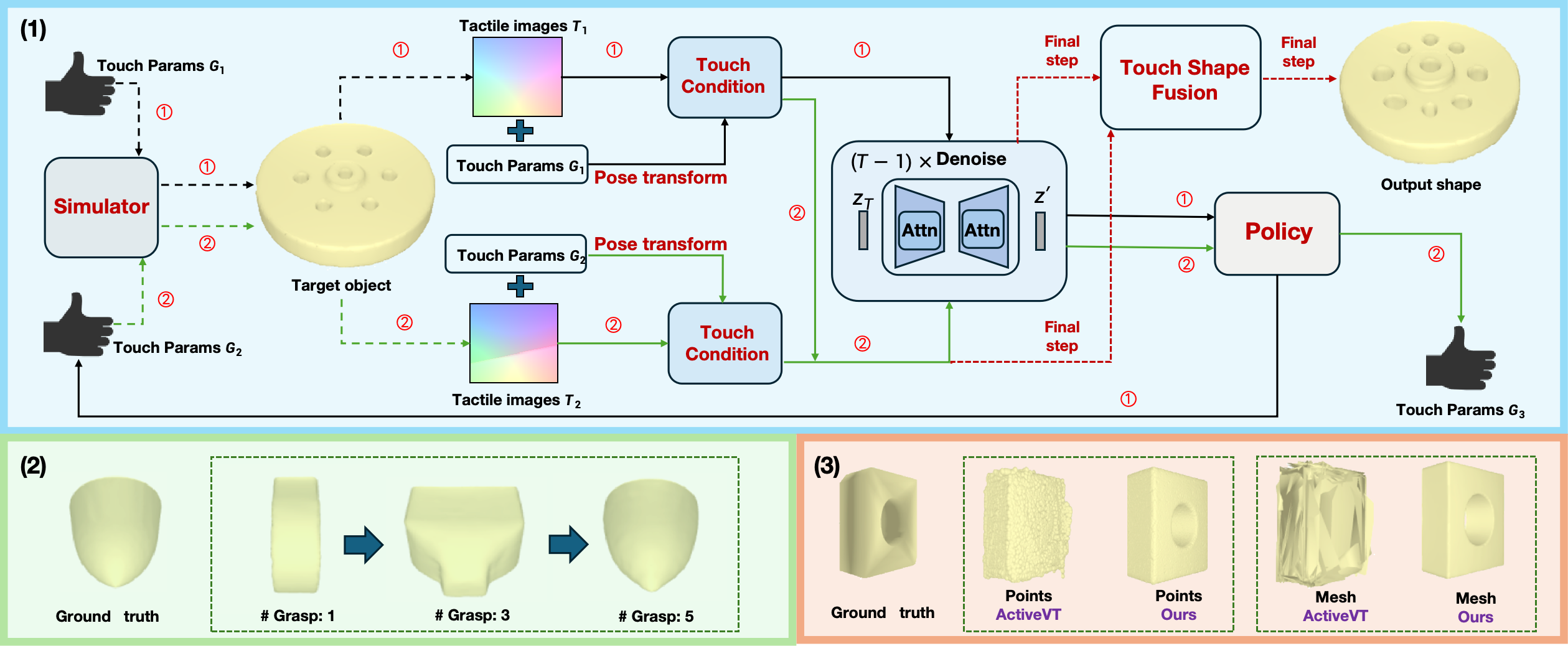}
    \captionof{figure}{  \label{fig:fig_teaser}
    (1) Exploring the target object and capturing the tactile image to reconstruct the 3D shape. We trained a diffusion model to obtain a low-dimensional and compact latent vector, which is used for predicting next touch location and reconstructing the target shapes. The numbers \normalsize{\textcircled{\scriptsize{1}}}\normalsize \ with black arrows  and \normalsize{\textcircled{\scriptsize{2}}}\normalsize \ with greeen arrows in the diagram represent two consecutive time steps. We only generate the full reconstruction at the final step. (2) As the touch exploration progresses, the reconstruction results gradually approach the ground truth. (3) Reconstruction results compared with ActiveVT \cite{activevt} (under visual-tactile settings).
    } 
\end{center}
}]

\begin{abstract}

Diffusion models have made breakthroughs in 3D generation tasks. Current 3D diffusion models focus on reconstructing target shape from images or a set of partial observations. While excelling in global context understanding, they struggle to capture the local details of complex shapes and limited to the occlusion and lighting conditions. To overcome these limitations, we utilize tactile images to capture the local 3D information and propose a Touch2Shape model, which leverages a touch-conditioned diffusion model to explore and reconstruct the target shape from touch. For shape reconstruction, we have developed a touch embedding module to condition the diffusion model in creating a compact representation and a touch shape fusion module to refine the reconstructed shape. For shape exploration, we combine the diffusion model with reinforcement learning to train a policy. This involves using the generated latent vector from the diffusion model to guide the touch exploration policy training through a novel reward design. Experiments validate the reconstruction quality thorough both qualitatively and quantitative analysis, and our touch exploration policy further boosts reconstruction performance.

\end{abstract}   
%---------------------------------------------------------------------------------
\footnotetext[1]{Corresponding authors.}
%---------------------------------------------------------------------------------
\section{Introduction}
\label{sec:intro}

\indent 3D generation and reconstruction tasks have emerged as key focal points in the fields of computer vision and graphics, offering valuable applications in areas like autonomous driving and robot interactions characterized by environmental occlusions and camera measurement errors \cite{diffusionsdf, ic3d, sdfusion}. However, acquiring 3D data presents greater challenges and costs compared to 2D image and text data. This underscores the critical importance of ongoing research into 3D reconstruction and generation.

Diffusion models have garnered substantial attention in generative tasks for their innovative approaches and promising results. While their impact has been particularly notable in 2D image generation \cite{diffimagegen1,diffimagegen3}, these models are also being applied to 3D tasks. Pioneering projects such as SDFusion \cite{sdfusion} and DiffusionSDF \cite{diffusionsdf} have demonstrated the ability to create the 3D shapes using visual images or multi-modal data. However, the current focus is on 3D shape reconstruction based on predetermined partial observations. This limitation presents challenges on two fronts. Firstly, while current methods can estimate the overall shape of a target from limited data, they may overlook the local details of objects, making it challenging to generate complex shapes. This motivates further research into 3D reconstruction across new modalities. Secondly, practical scenarios present challenges where environmental factors like occlusions and varying lighting conditions impede the complete extraction of crucial local information. The complexity of real-world conditions underscores the necessity of actively exploring the target to capture information and facilitate the reconstruction process.

In this work, we employ a simulated robotic arm guided by a trained policy model to touch the target, enabling the acquisition of tactile images to reconstruct the target through touch interaction. Tactile images provide local 3D shape information, including spatial contact points and detailed shape information, enhancing the reconstruction of local details, while visual data assists in predicting overall shape characteristics. Especially, we propose Touch2Shape, a touch-conditioned diffusion model for shape exploration through continuous touch and shape reconstruction by integrating all captured tactile images (along with the optional visual image). The reason for utilizing the diffusion model is to leverage its powerful generative capabilities to assist the model in generating a latent shape when the initial data perception is limited. The latent shape is then used to guide the policy model to predict beneficial touch locations for reconstruction based on potential missing areas. To train the exploration policy, we integrate the diffusion model with reinforcement learning and design a reward function. 

In contrast to previous methods, our approach not only produces superior 3D reconstruction outputs but also eliminates the need to generate the final shape at every step. Due to the fact that the diffusion model can generate a low-dimensional and compact latent space, we only generate the full reconstruction of the target object at the final step. Extensive experiments validate the effectiveness of our method, demonstrating significant improvements in both reconstruction performance and the ability to improve reconstruction quality through touch exploration.

The main contributions of this article are as follows:
\begin{itemize}
\item{We propose Touch2Shape, a touch-conditioned 3D diffusion model for shape exploration and reconstruction, utilizing the latent vector to guide the touch location planning and shape decoding.}
\item{Touch2Shape utilizes contrastive touch encoder to embed touch information and shapes in a joint space.}
\item{We propose a touch shape fusion module to optimize the reconstructed shape using touch information.}
\item{We combine the diffusion model with reinforcement learning for shape exploration policy training and design the corresponding reward function.}
\end{itemize}
\section{Related Work}
\label{sec:related}

% \subsection{Touch-based 3D Object Reconstruction}
% \label{sec:related:2.1}

\textbf{Touch-based 3D Object Reconstruction.} There are a lot of works addressing 3D shape reconstruction from visual signals. The approaches in this field vary depending on the type of visual input utilized, such as single view RGB images \cite{single}, multi-view RGB images \cite{multi}, and depth images \cite{fcs1}. Additionally, the types of 3D representations they predict also differ, encompassing aspects such as voxels \cite{vi-voxel}, point clouds \cite{point}, meshes \cite{mesh}, and signed distance functions \cite{sdf}. However, due to the high-dimensionality of the observation space and the interference of occlusions and external lighting conditions, training computer vision algorithms for object manipulation faces significant challenges. Recently, some researchers use high-resolution tactile perception such as DIGIT \cite{digit} and Gelsight sensor \cite{gelsight} to accomplish 3D object reconstruction. This type of perception, compared to vision-based perception, can obtain rich contact information and remains effective in the face of occlusions, transparent or reflective object materials. TouchSDF \cite{touchsdf} predicts the local 3D shape corresponding to the tactile image and completes the 3D expression of the object by training an implicit neural function to represent the signed distance. Smith et al. \cite{vtrecon} propose the first approach for reconstruction using both vision and touch, highlighting their complementary nature. Combining vision and touch modalities can yield richer information and improve 3D shape reconstruction. In this work, our Touch2Shape model trains a touch-conditioned diffusion model for 3D shape reconstruction, supporting settings with both tactile only and visual-tactile inputs.

%-------------------------------------------------------------------------

%------------------------------------------------------------------------
\begin{figure*}[tp!]
  \centering
  \includegraphics[width=0.999\linewidth]{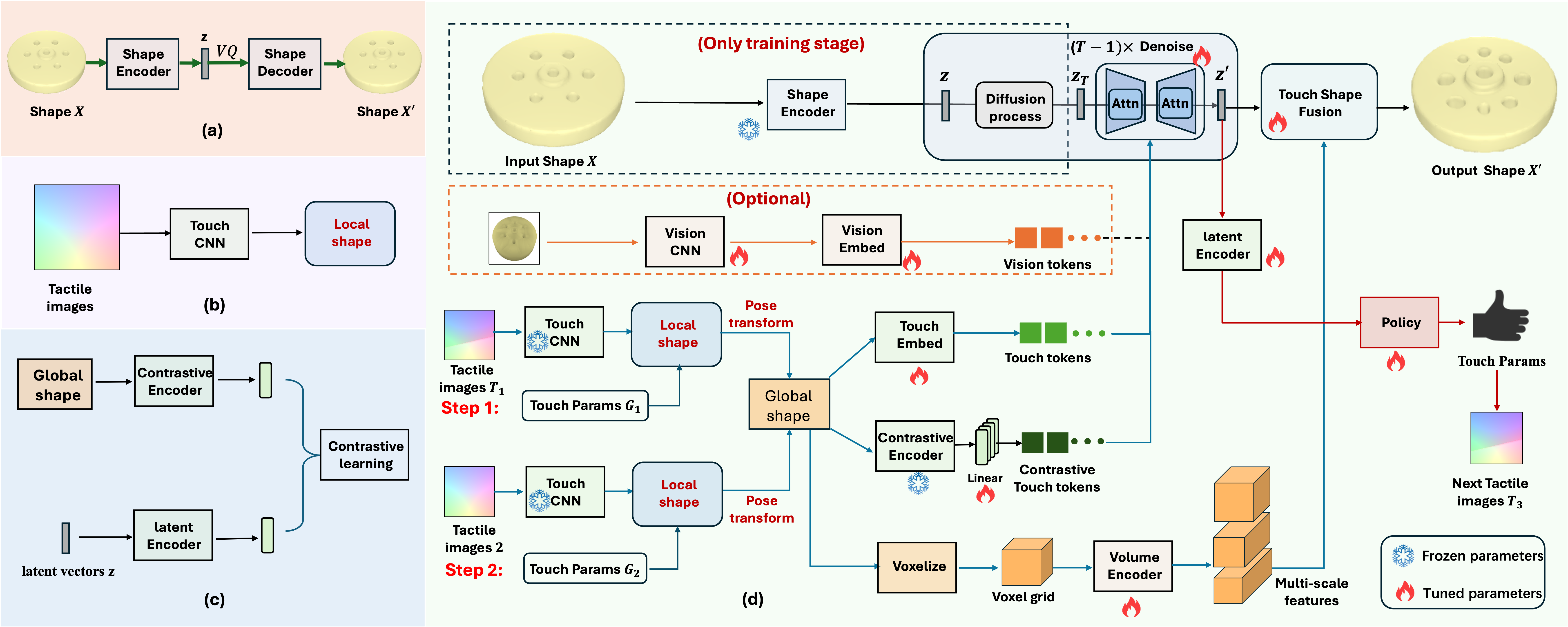}
  \caption{\label{fig:fig_pipeline}
          We pretrained (a) the shape encoder and decoder, (b) the touch CNN model that is used for touch chart prediction, and (c) the contrastive touch encoder. During the training of (d) the touch-conditioned 3D diffusion model, we kept the parameters of the pretrained modules fixed and train a touch-conditioned 3D diffusion model for touch exploration policy learning and shape reconstruction.
          }
% \vspace{-0.2cm} 
\end{figure*}

% \subsection{3D Diffusion Model}
% \label{sec:related:2.2}

\textbf{3D Diffusion Model.} The diffusion probability model \cite{diffusion1,diffusion2} operates by gradually removing noise from data points to generate samples from a distribution. The diffusion model in the field of images has garnered widespread attention, for tasks such as image generation \cite{diffimagegen1,diffimagegen3}, super-resolution \cite{diffimagesr2}, image-editing \cite{diffimageedit2}.The application of the diffusion models extends well into 3D generation tasks. Some works train diffusion models to generate point clouds \cite{diff3dpoint1,diff3dpoint2}, voxel grids \cite{diff3dvoxel2,ic3d} and meshes \cite{diff3dmesh}. Recently, SDFusion \cite{sdfusion} encodes 3D shapes into an expressive low-dimensional latent space, which is used to train the 3D diffusion model and present the shape as a TSDF (truncated signed distance function) volume. DiffusionSDF \cite{diffusionsdf} combine a 3D diffusion model with implicit expressions to represent the target 3D surface using a neural symbolic distance function and generate diverse reconstruction results using a diffusion model. These methods excel in both unconditional generation and conditional generation from partial inputs, inspiring our study of touch-based shape reconstruction tasks using the diffusion model.

%-------------------------------------------------------------------------

% \subsection{Shape Exploration}
% \label{sec:related:2.3}

\textbf{Shape Exploration.} In this paper, the purpose of shape exploration is to actively capture continuous tactile information and enhance shape reconstruction accuracy based on these acquired tactile data. One approach to achieve this goal is to maximize the information from the captured tactile data such as Monte Carlo sampling \cite{infogain1} and Gaussian Process Regression \cite{infogain2}. In recent years, many strategies have incorporated deep learning to perform active sensing mechanisms in shape exploration. Most methods \cite{ijcai18, activevision1, activevision2, activevision4, zhaoxuan1, zhaoxuan2, zhaoxuan3} have been developed for the active vision task by planning camera perspectives to capture multi-view images essential for 3D reconstruction. For active touch sensing, several prior works \cite{activevh, activetouch1, activetouch2, weiboyan, activetouch3, activetouch4} address the problem of shape reconstruction by estimating uncertainty for the selection of the next touch. In this paper, we combine the diffusion model with reinforcement learning to learn the exploration policy. The most relevant work is \cite{ijcai18} and \cite{activevt}. The former proposes a unified model for view planning and vision-based object reconstruction, which combines the 3D reconstruction learning and reinforcement learning. The latter proposes a mesh-based 3D shape reconstruction model where touch exploration strategies are learned over shape predictions using reinforcement learning. In contrast to existing methodologies, our study focuses on training a touch-based diffusion model to generate a latent space, obviating the need to construct a complete shape at every step. Additionally, we leverage the captured touch information to enhance the final shape output. These approaches enable our method to achieve increased reconstruction quality improvements with continuous touches. 

%-------------------------------------------------------------------------

\section{Method}
\label{sec:method}

As shown in Figure \ref{fig:fig_teaser}, we train a touch-conditioned diffusion model to implicitly represent the target object information for shape exploration and reconstruction. In test stage, we gather tactile images $(T_0,...,T_{n-1})$ from the target, utilizing the trained diffusion model to obtain a low-dimensional representation for predicting the subsequent touch location. The shape is also reconstructed based on all the captured tactile information.

Figure \ref{fig:fig_pipeline} illustrates the architecture of our Touch2Shape Model. Following SDFusion \cite{sdfusion}, we employ the volumetric Truncated Signed Distance Field (T-SDF) to model the distribution across 3D shapes and a 3D variant of the Vector Quantised-Variational AutoEncoder (VQ-VAE) \cite{vqvae} for encoding and decoding 3D shapes. Let the input shape is represented as the T-SDF volume $X \in R^{D \times D \times D}$, the encoder is $E_s$ and the decoder is $D_s$, we encode the input shape into the latent vector $z$ and decode the vector to the generated shape $X'$:
\begin{equation}\label{eqn1}
z = E_{s}(X) \text{,} \ \ \ \text{and} \ \ \ X' = D_{s}(VQ(z))
\text{,}
\end{equation}
where $VQ$ is the quantization step.

We pretrained the VQ-VAE model on both ABC dataset \cite{abc} and ShapeNet dataset \cite{shapenet}. Using the encoded latent vector, gaussian noise is added at random timestaps $t$, and a denoising model is trained based on the touch condition (Section \ref{sec:method:3.1}). The touch charts prediction model is pretrained as \cite{vtrecon, activevt}. Contrastive learning is employed to embed touch information and shapes in a joint space. The policy model receives the denoised vector as input and is trained using reinforcement learning (Section \ref{sec:method:3.2}). The generation of the final shape is optimized through tactile data (Section \ref{sec:method:3.3}).

%-------------------------------------------------------------------------

\subsection{Touch-conditioned Diffusion Model}
\label{sec:method:3.1}

Given a sample $z$ encoded by the pretrained shape encoder $E_s$, we learn a touch-conditioned distribution for touch based shape exploration and reconstruction. The loss function for diffusion model training is as follows:
\begin{equation}\label{eqn2}
L_{diff}(t,n) = ||{E}_{\theta}(z_t, r(t), C(T_0,...,T_{n-1})) - {\epsilon}_{t} ||_2
\text{,}
\end{equation}
where ${\epsilon}_{t}$ is the added gaussian noise, the ${E}_{\theta}$ is the denoising network, $C$ is the touch condition extraction network for $n$ touch inputs $(T_0,...,T_{n-1})$, $t$ is the number of diffusion timesteps and $r(t)$ is time embedding respectively.

\textbf{Touch Embedding.} 
We pretrain a TouchCNN model as \cite{vtrecon, activevt} to predict the touch charts. We assume that we can obtain up to N tactile images, where each tactile image generates a chart as a tensor of size $M \times 4$ (each vertex contains coordinates $x$, $y$, $z$, and touch status, M is the number of chart vertices). By merging all the vertices of these charts together, we create a tensor of size $N \times M \times 4$. If we have fewer than $N$ tactile images, the coordinates of the vertices for the remaining charts are zeroed out. Each touch chart is considered as a token. We first apply position encoding to the centroid of each chart, then perform convolution operations on the merged tensor to extract vertex features. After pooling operations and adding position embedding, we finally obtain $N$ tokens.

\textbf{Contrastive Touch Encoder.} It has been confirmed in IC3D \cite{ic3d} that the joint encoding of both 2D and 3D information is beneficial to condition the generation of 3D shapes based on images. UniTouch \cite{unitouch} connects the tactile signals to other modalities, including vision, language, and
sound. In this study, we propose a contrastive model for embedding touch and latent vectors in a joint space. Utilizing the latent vectors generated by the pretrained encoder $E_s$, we establish a latent encoder to acquire the shape features.
For touch features, we construct a network similar to the touch embedding model mentioned above, except for the addition of a pooling layer at the end. We utilize moco \cite{moco} for training the contrastive learning task. We set touch features as queries and shape features as keys. The objective is to pulling together with matching touch-shape pairs while pushing unmatched pairs apart. The loss function is:
\begin{equation}\label{eqn3}
L_{cl} = -log \frac{e^{q \cdot k_{p}/\tau}}{\sum_{i=0}^{K} e^{q \cdot k_i/\tau}}
\text{,}
\end{equation}
where $q$ is the query feature, $k$ is the key feature, $k_i$ is the i-th element in the queue of size K, and $\tau$ is the temperature parameter respectively.

\textbf{Visual-tactile Setting.} Touch2Shape model supports two modes: tactile only and visual-tactile. In the visual-tactile setting, due to the ability of visual information to capture the global context, reasonable shape outputs can be obtained even when initial tactile images are limited. We initially use visual information to guide tactile exploration, predict the global structure, and refine the shape as more tactile information becomes available. The implementation involves extracting feature tokens from images using ResNet \cite{resnet}, combining them with touch tokens through a dropout layer, and then inputting them together into the denoising network of the Diffusion model. This process generates denoised vectors $z'$, which is then used for 
shape reconstruction and next touch location prediction.

%-------------------------------------------------------------------------

\subsection{Touch Shape Fusion}
\label{sec:method:3.2}

\begin{figure}[htbp]
  \centering
  \includegraphics[width=0.999\linewidth]{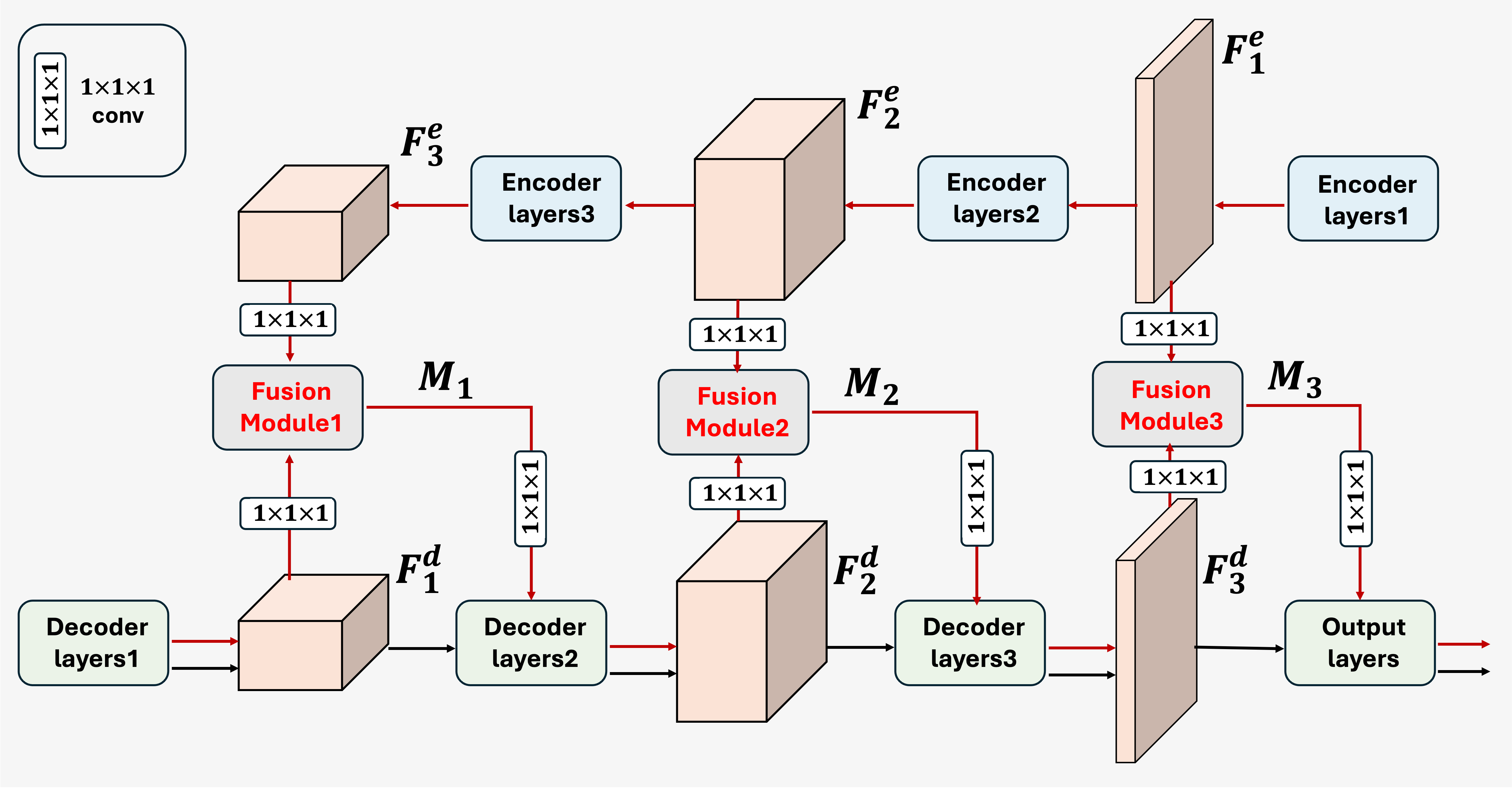}
  \caption{\label{fig:fig_network}
         Touch shape fusion module. The black arrows indicate the flow of the shape decoder, while the red arrows represent the flow after incorporating the touch shape fusion module. To simplify, we use 2D grids to visualize the 3D feature maps.
         }
% \vspace{-0.4cm} 
\end{figure}

The touch shape fusion module is designed with two goals. Firstly, the diffusion model typically generates shapes that are globally consistent with the input conditions, but may exhibit discrepancies in local details against the input touch chart. Secondly, the encoded latent vector, being low-dimensional and compressed, risks omitting crucial high-dimensional details. To address this, we voxelize the global shape merged from all historical touch information (as Figure \ref{fig:fig_pipeline}) and utilize an additional voxel encoder to capture multi-scale features. These features are fused with the different scale features generated during the decoding of the latent vector, resulting in a better shape reconstruction. The decoder originates from the pre-trained VQVAE model and is fine-tuned in touch shape fusion module training. Figure \ref{fig:fig_network} depicts the network module. We represent the feature at location $(c,k,j,i)$ with a size of $C \times D \times H \times W$ is $M(c,k,j,i)$. Taking the first fusion block as example, the fusion feature $M_{1}(c,k,j,i)$ can be computed as follows:
\begin{equation}\label{eqn4}
M_1(c,k,j,i) = \frac{\lambda \cdot F_{3}^{e}(c,k,j,i) \cdot e^{F_{1}^{d}(c,k,j,i)}}{\sum_{c'=0}^{c'=C-1}(e^{F_{1}^{d}(c',k,j,i)})}
\text{,}
\end{equation}
where $F_{1}^{d}$ and $F_{3}^{e}$ is the 3D feature maps as depicted in Figure \ref{fig:fig_network}, $\lambda$ is a learnable weight.

%------------------------------------------------------------------------

%-------------------------------------------------------------------------

\subsection{Policy Training}
\label{sec:method:3.3}

This module aims to create potential 3D shapes utilizing the diffusion model based on the acquired tactile (and optional visual) data, which are then employed to determine the next touch location for shape exploration. Previous methods \cite{activevt} required generating the final shape at each time step and encoding the shape into a latent vector, or directly predicting touch location parameters from the generated shape using a policy network. The reward setting was based on the change in Chamfer Distance (CD) \cite{cd} after two consecutive touches. In this paper, we combine the diffusion model with policy training. At each time step, we input the latent vector $z$ of the target object, add noise through the diffusion model, and then use a touch-conditioned denoising network to obtain a denoised latent vector $z'$. This vector is a low-dimensional 3D feature map. We first employ the pre-trained latent encoder in Figure \ref{fig:fig_pipeline} (c) to encode both the initial and current latent vectors of the touch-conditioned diffusion model. Subsequently, we construct an action embedding module to derive the embedding for the potential actions. Each action is identified by its positional index on a sphere consisting of 50 actions, as described in \cite{activevt}. By combining these three vectors, we employ fully connected layers to predict the value associated with each action.

The complete process obviates the necessity of predicting the final high-dimensional T-SDF Volume. Instead, we utilize the shape decoder and shape fusion only at the final time step, thereby achieving a separation of shape decoder and shape exploration. For reward function setting, since the final output shape is not predicted, we design it to be the difference in the diffusion model's loss values. The reward at step $n$ is computed as follows:
\begin{equation}\label{eqn5}
R = H(L_{diff}(t,n-1) - L_{diff}(t,n))
\text{,}
\end{equation}
where $H(\cdot)$ assigns a value of 1 to all values greater than or equal to 0, and a value of 0 to those less than 0.

It can be observed that our objective is to encourage actions that help the touch-conditioned diffusion model generate a denoised vector z' closer to the latent vector z. It can be trained by reinforcement learning such as DQN \cite{dqn}. The loss function for reinforcement learning is as follows:
\begin{equation}\label{eqn6}
\begin{aligned}
L_{rl} = [R + \gamma \max_{a_{n+1}} Q(T_0,...,T_{n+1}, a_{n+1}) \\ - Q(T_0,...,T_{n}, a_{n})] ^2
\text{,}
\end{aligned}
\end{equation}
where $Q(\cdot)$ is the Q-value function approximated by the network, $a_n$ is the  action taken on the timestep n, and $\gamma$ is the discount factor.

%-------------------------------------------------------------------------

%%-----------------------------------------------------------------------
\begin{table}[htbp]
\renewcommand{\arraystretch}{1.0}
\begin{center}
\begin{tabular}{|  c | c |  c  c |}
\hline
\multirow{2}{*}{\makecell{\textbf{\makebox[0.12\textwidth][c]{Method}}}} &\multirow{2}{*}{\makecell{\textbf{\makebox[0.1\textwidth][c]{Mode}}}}  &\multicolumn{2}{c|}{\textbf{{Grasp \#}}} \\
\cline{3-4}
   &   &\makebox[0.08\textwidth][c]{0}  &\makebox[0.08\textwidth][c]{1}  \\
\hline
VTRecon  &\multirow{3}{*}{\makecell{\textbf{{T}}}} &25.586 &9.016  \\
ActiveVT  &  &24.864 &8.220 \\
Ours  &    &40.283   &\textbf{6.794} \\
\hline
VTRecon  &\multirow{3}{*}{\makecell{\textbf{{T+V}}}} &2.653 &2.637  \\
ActiveVT  &  &2.538 &2.486 \\
Ours &   &1.475 &\textbf{1.406} \\
\hline
\end{tabular}
\caption{ \label{tab_abc}
    Experimental results for different settings and different numbers of grasps on dataset ABC. The evaluation metric is CD (lower is better). 
} 
\end{center}
\vspace{-0.4cm} 
\end{table}
%-------------------------------------------------------------------------

%%-----------------------------------------------------------------------
\begin{table*}[tp]
\renewcommand{\arraystretch}{1.0}
\begin{center}
\begin{tabular}{|  c | c |  c |  c | c | c | c | c | c | c |}
\hline
\multirow{3}{*}{\makecell{\textbf{\makebox[0.055\textwidth][c]{Category}}}}  &\multicolumn{9}{c|}{\textbf{\# Touches (Unseen objects and poses)}} \\
\cline{2-10}
 &\multicolumn{3}{c|}{\textbf{1}}  &\multicolumn{3}{c|}{\textbf{10}}  &\multicolumn{3}{c|}{\textbf{20}}      \\
\cline{2-10}   
   &\makebox[0.075\textwidth][c]{TouchSDF} &\makebox[0.075\textwidth][c]{Ours$_{T}$} &\makebox[0.075\textwidth][c]{Ours$_{TV}$} &\makebox[0.075\textwidth][c]{TouchSDF} &\makebox[0.075\textwidth][c]{Ours$_{T}$} &\makebox[0.075\textwidth][c]{Ours$_{TV}$} &\makebox[0.075\textwidth][c]{TouchSDF} &\makebox[0.075\textwidth][c]{Ours$_{T}$} &\makebox[0.075\textwidth][c]{Ours$_{TV}$} \\
\hline
Bottle &0.113 &0.110 &0.036 &0.082 &0.042 &0.034 &0.047 &0.041 &\textbf{0.033}  \\
Mug &0.091 &0.118 &0.042 &0.072 &0.050 &0.042 &0.066 &0.049 &\textbf{0.042} \\
Bowl &0.085 &0.128 &0.046 &0.073 &0.051 &0.045 &0.048 &0.049 &\textbf{0.040} \\
Camera &0.131 &0.132 &0.055 &0.101 &0.058 &0.052 &0.092 &0.056 &\textbf{0.050} \\
Guitar &0.195 &0.139 &0.059 &0.177 &0.067 &0.056 &0.155 &0.064 &\textbf{0.045} \\
Jar &0.164 &0.122 &0.048 &0.136 &0.059 &0.046 &0.071 &0.055 &\textbf{0.045} \\
\hline
Average &0.136 &0.124 &0.048 &0.112 &0.056 &0.046 &0.081 &0.053 &\textbf{0.042} \\
\hline
\end{tabular}
\caption{\label{tab_shapenet}
    Experimental results for different numbers of touches on dataset ShapeNet. 
     Ours$_{T}$ and Ours$_{TV}$ respectively represent our methods under the tactile only and visual-tactile settings. The evaluation metric is EMD (lower is better). 
}
\end{center}
% \vspace{-0.4cm} 
\end{table*}
%-------------------------------------------------------------------------

\section{Experiment}
\label{sec:exp}

In this section, we describe the experiment settings and then compare our model with the state-of-art touch-based 3D reconstruction methods and validate our policy training strategy. Through the ablation study, we validate the necessity of each module. Additional experimental details can be found in the supplementary material. 

\subsection{Experimental Settings}
\label{sec:exp1}

\textbf{Datasets.} We validate our model using two datasets. The first dataset utilized is derived from \cite{vtrecon, activevt}, built upon the ABC dataset \cite{abc}. This dataset comprises 40,000 objects with ambiguous class definitions and diverse shapes, presenting a significant generalization hurdle. The second dataset employed originates from \cite{touchsdf}, encompassing 1650 ShapeNet \cite{shapenet} objects that span six categories: bowls, bottles, cameras, jars, guitars, and mugs. All the touch images and initial RGB images are rendered using the simulation environment introduced in \cite{activevt}.

\textbf{Training Setup.} We pre-trained the VQ-VAE model following SDFuison \cite{sdfusion} and the touch chart prediction model following \cite{vtrecon, activevt}. The volume resolution is set at 64x64x64. The training process is segmented into three parts: diffusion model training, touch shape fusion module training, and policy training. The diffusion model and touch shape fusion can be trained concurrently since they do not share any components. The diffusion model was trained for 1 million iterations with an initial learning rate of 0.00001 and batch size of 12, while the touch shape fusion was trained for 250,000 iterations with an initial learning rate of 0.0001 and batch size of 8. After the diffusion model training finished, we conducted policy training in silmulation environment \cite{activevt} for 200 epochs with a learning rate of 0.0003 and batch size of 16. All the network codes are implemented using the PyTorch framework \cite{pytorch} and trained on the GeForce RTX 4090 for approximately a week.

\textbf{Evaluation Metrics.} Following previous learning-based reconstruction tasks, we use Chamfer Distance (CD) \cite{cd} and Earth Mover’s Distance (EMD) \cite{emd} for reconstruction evaluation. The former is a common 3D reconstruction metric for measuring the point-wise distance between two pointsets. The latter is a metric used to measure the dissimilarity that calculates the minimum amount of work needed to transform one point set into another, which is particularly useful in comparing the visual quality of 3D shapes.

%-------------------------------------------------------------------------

\subsection{Evaluation on Reconstruction Performance}
\label{sec:exp2}

\begin{figure*}[tp!]
\begin{center}
  \begin{minipage}{0.12\linewidth}
  \centering{\textbf{Ground \\ truth}}
  \end{minipage}
  \begin{minipage}{0.14\linewidth}
  \centerline{\includegraphics[width=1\textwidth, height=0.8\textwidth]{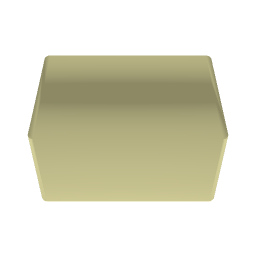}}
  \end{minipage}
  \begin{minipage}{0.14\linewidth}
  \centerline{\includegraphics[width=1\textwidth, height=0.8\textwidth]{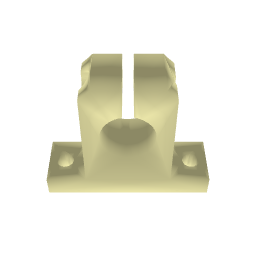}}
  \end{minipage}
  \begin{minipage}{0.14\linewidth}
  \centerline{\includegraphics[width=1\textwidth, height=0.8\textwidth]{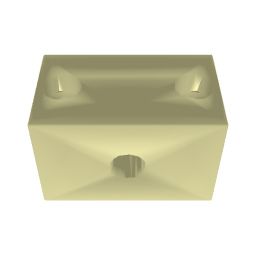}}
  \end{minipage}
  \begin{minipage}{0.14\linewidth}
  \centerline{\includegraphics[width=1\textwidth, height=0.8\textwidth]{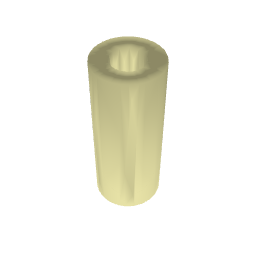}}
  \end{minipage}  
  \begin{minipage}{0.14\linewidth}
  \centerline{\includegraphics[width=1\textwidth, height=0.8\textwidth]{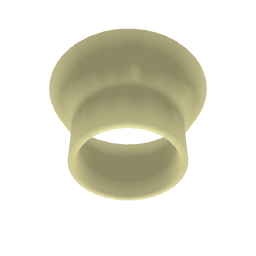}}
  \end{minipage}
  \begin{minipage}{0.14\linewidth}
  \centerline{\includegraphics[width=1\textwidth, height=0.8\textwidth]{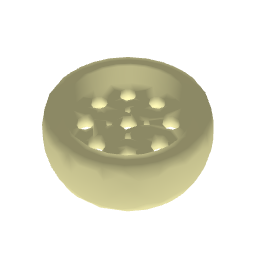}}
  \end{minipage}

  \begin{minipage}{0.12\linewidth}
  \centering{\textbf{points \\ (ActiveVT)}}
  \end{minipage}
  \begin{minipage}{0.14\linewidth}
  \centerline{\includegraphics[width=1\textwidth, height=0.8\textwidth]{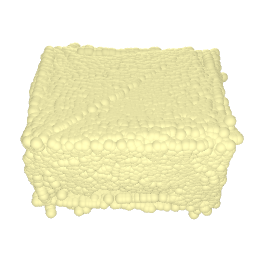}}
  \end{minipage}
  \begin{minipage}{0.14\linewidth}
  \centerline{\includegraphics[width=1\textwidth, height=0.8\textwidth]{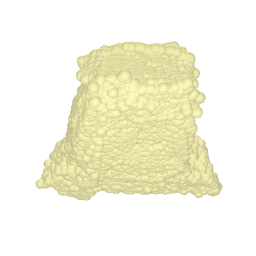}}
  \end{minipage}
  \begin{minipage}{0.14\linewidth}
  \centerline{\includegraphics[width=1\textwidth, height=0.8\textwidth]{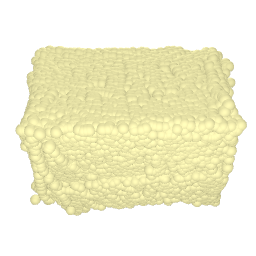}}
  \end{minipage}
  \begin{minipage}{0.14\linewidth}
  \centerline{\includegraphics[width=1\textwidth, height=0.8\textwidth]{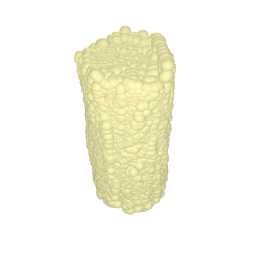}}
  \end{minipage}  
  \begin{minipage}{0.14\linewidth}
  \centerline{\includegraphics[width=1\textwidth, height=0.8\textwidth]{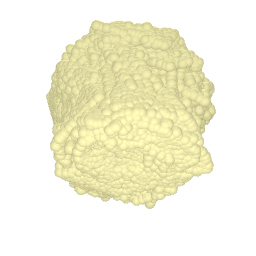}}
  \end{minipage}
  \begin{minipage}{0.14\linewidth}
  \centerline{\includegraphics[width=1\textwidth, height=0.8\textwidth]{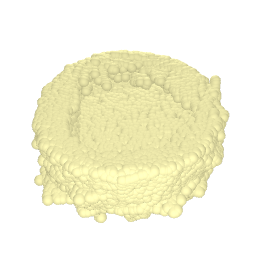}}
  \end{minipage}  

  \begin{minipage}{0.12\linewidth}
  \centering{\textbf{points \\ (ours)}}
  \end{minipage}
  \begin{minipage}{0.14\linewidth}
  \centerline{\includegraphics[width=1\textwidth, height=0.8\textwidth]{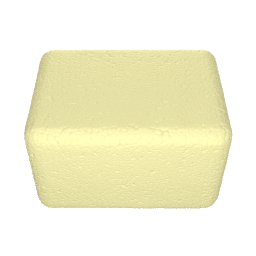}}
  \end{minipage}
  \begin{minipage}{0.14\linewidth}
  \centerline{\includegraphics[width=1\textwidth, height=0.8\textwidth]{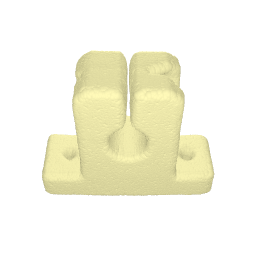}}
  \end{minipage}
  \begin{minipage}{0.14\linewidth}
  \centerline{\includegraphics[width=1\textwidth, height=0.8\textwidth]{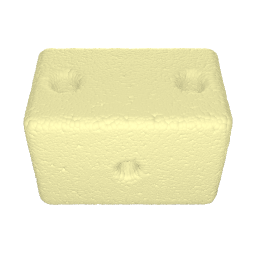}}
  \end{minipage}
  \begin{minipage}{0.14\linewidth}
  \centerline{\includegraphics[width=1\textwidth, height=0.8\textwidth]{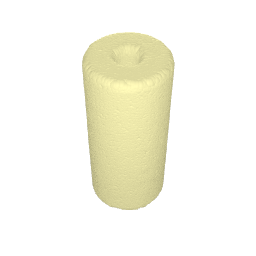}}
  \end{minipage}  
  \begin{minipage}{0.14\linewidth}
  \centerline{\includegraphics[width=1\textwidth, height=0.8\textwidth]{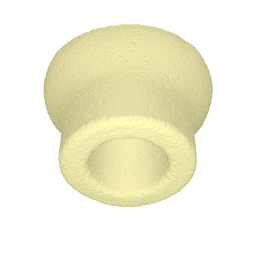}}
  \end{minipage}
  \begin{minipage}{0.14\linewidth}
  \centerline{\includegraphics[width=1\textwidth, height=0.8\textwidth]{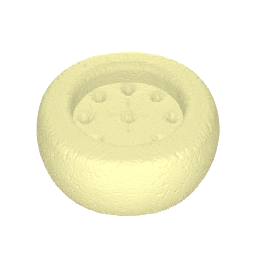}}
  \end{minipage}  

  \begin{minipage}{0.12\linewidth}
  \centering{\textbf{mesh \\ (ActiveVT)}}
  \end{minipage}
  \begin{minipage}{0.14\linewidth}
  \centerline{\includegraphics[width=1\textwidth, height=0.8\textwidth]{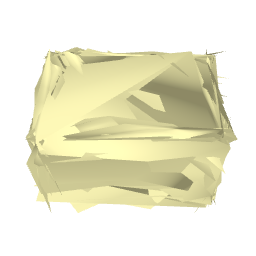}}
  \end{minipage}
  \begin{minipage}{0.14\linewidth}
  \centerline{\includegraphics[width=1\textwidth, height=0.8\textwidth]{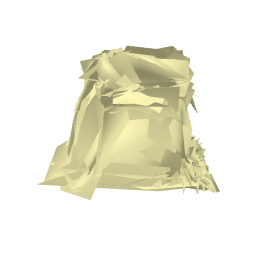}}
  \end{minipage}
  \begin{minipage}{0.14\linewidth}
  \centerline{\includegraphics[width=1\textwidth, height=0.8\textwidth]{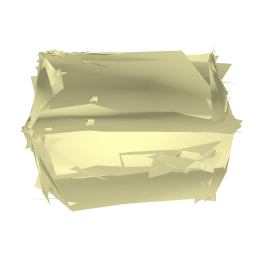}}
  \end{minipage}
  \begin{minipage}{0.14\linewidth}
  \centerline{\includegraphics[width=1\textwidth, height=0.8\textwidth]{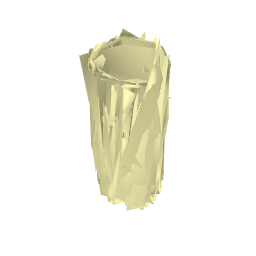}}
  \end{minipage}  
  \begin{minipage}{0.14\linewidth}
  \centerline{\includegraphics[width=1\textwidth, height=0.8\textwidth]{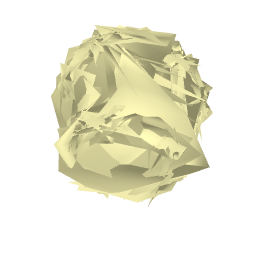}}
  \end{minipage}
  \begin{minipage}{0.14\linewidth}
  \centerline{\includegraphics[width=1\textwidth, height=0.8\textwidth]{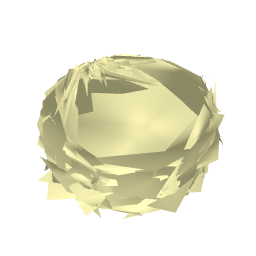}}
  \end{minipage}  

  \begin{minipage}{0.12\linewidth}
  \centering{\textbf{mesh \\ (ours)}}
  \end{minipage}
  \begin{minipage}{0.14\linewidth}
  \centerline{\includegraphics[width=1\textwidth, height=0.8\textwidth]{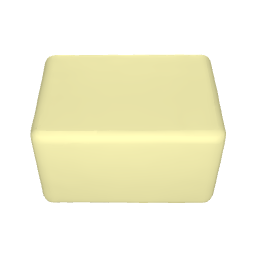}}
  \end{minipage}
  \begin{minipage}{0.14\linewidth}
  \centerline{\includegraphics[width=1\textwidth, height=0.8\textwidth]{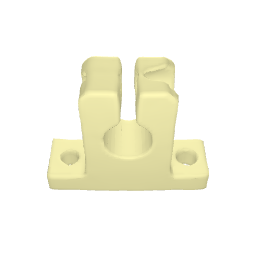}}
  \end{minipage}
  \begin{minipage}{0.14\linewidth}
  \centerline{\includegraphics[width=1\textwidth, height=0.8\textwidth]{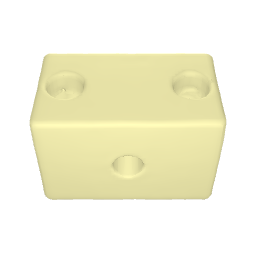}}
  \end{minipage}
  \begin{minipage}{0.14\linewidth}
  \centerline{\includegraphics[width=1\textwidth, height=0.8\textwidth]{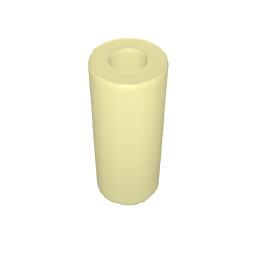}}
  \end{minipage}  
  \begin{minipage}{0.14\linewidth}
  \centerline{\includegraphics[width=1\textwidth, height=0.8\textwidth]{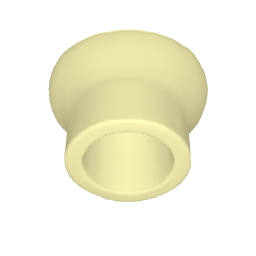}}
  \end{minipage}
  \begin{minipage}{0.14\linewidth}
  \centerline{\includegraphics[width=1\textwidth, height=0.8\textwidth]{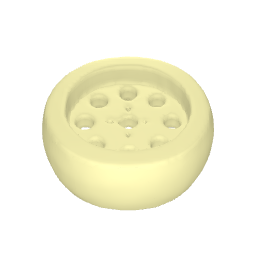}}
  \end{minipage}

  \caption{ \label{fig:fig_sdfvt}
           Qualitative results of ActiveVT \cite{activevt} and ours. While ActiveVT struggles with visualizations and detail preservation, our method excels in maintaining global shape accuracy across diverse structures, ensuring satisfactory local details.}
%  976, 649, 661, 874, 1616, 2752, 908, 4136
% failure: 958, 1011, 1037, 1046, 1264, 1480, 2624
\end{center}
% \vspace{-0.4cm} 
\end{figure*}

\textbf{Results on the ABC dataset.} we evaluate our reconstruction model and compare it with \cite{vtrecon} and \cite{activevt}. The former method (we called VTRecon here) predicts a local chart at each touch site and combines them with vision signals to predict global charts via graph convolutional networks. The latter method (we called ActiveVT here) proposes an active touch sensing for 3D reconstruction method to improve the reconstruction performance. We examine the performance across 2 settings: (1) tactile only (T), only touch signals from all hand sensors are used during shape exploration, the hand has 4 fingers thus we can obtain at most 4 valid tactile images during one touch exploration. (2) touch and vision (T+V), an extension of tactile only setting with adding an initial RGB image of the target object. To calculate the Chamfer Distance (CD) for our SDF volume, we run marching cubes to get the object meshes and extract 30,000 points from each sample. We run the evaluation script 5 times and calculate the average results for report. As shown in Table \ref{tab_abc}, it is clear that when obtaining tactile image input, our method is superior to other methods. Especially on the visual-tactile 3D reconstruction task, we obtain a very low CD error, which validates the multi-modal fusion ability of our model. The visualization results of V+T settings (one grasp) are shown in Figure \ref{fig:fig_sdfvt}. ActiveVT generally produces poor visualizations on the mesh surfaces. For point cloud generation, ActiveVT can preserve shapes similar to the ground truth for some structurally simple objects, but for complex shapes or objects with cavities, it loses many details. In contrast, our method is capable of maintaining a complete global shape output for diverse shapes and provides a satisfactory result in local details as well. 

%----------------------------------------------------------------------------------

\textbf{Results on the ShapeNet dataset.} TouchSDF \cite{touchsdf} validates the reconstruction results across six categories in the ShapeNet dataset using the EMD metric. The dataset is devided into three subsets: 1,100 objects for training, 200 for validation and 350 for testing.  We evaluate our model using the same dataset partition and evaluation metric. For touch sensing, we adopt the setting of TouchSDF \cite{touchsdf}, which involves capturing tactile images by poking the target object, ensuring a fair comparison. Within this framework, we are limited to obtaining a maximum of one valid tactile image per touch action. The quantitative results are reported in Table \ref{tab_shapenet}. Note that TouchSDF only supports reconstruction from touch but our method supports both tactile only and visual-tactile settings. It can be observed that our method did not yield ideal results initially, but as the number of touches increased, our approach gradually surpassed TouchSDF, demonstrating our method's ability to integrate tactile information from different locations. Moreover, when combined with visual signals, our results were further enhanced, demonstrating the ability to integrate tactile and visual information to produce a better reconstruction. The visualization results on ShapeNet dataset are reported in the supplementary material.

%-------------------------------------------------------------------------

% -----------------------------------------------------------------------
\begin{table}[htbp]
\renewcommand{\arraystretch}{1.0}
\begin{center}
\begin{tabular}{|  c | c |  c  c  c c|}
 \hline
\makecell{\textbf{\makebox[0.047\textwidth][c]{Mode}}} &\makecell{\textbf{\makebox[0.047\textwidth][c]{Method}}}  &\makecell{\textbf{\makebox[0.049\textwidth][c]{Oracle}}}  &\makecell{\textbf{\makebox[0.049\textwidth][c]{Random}}} &\makecell{\textbf{\makebox[0.047\textwidth][c]{Even}}}  &\makecell{\textbf{\makebox[0.047\textwidth][c]{RL}}}  \\
\hline
\multirow{2}{*}{\makecell{\textbf{{T}}}} &ActiveVT  &16.38 &25.83 &24.53 &23.84  \\
&Ours &4.88 &8.14 &7.44 &6.63 \\
\hline
\multirow{2}{*}{\makecell{\textbf{{T+V}}}} &ActiveVT  &77.18 &90.65 & 90.29 & 89.32 \\
&Ours &75.01 &88.41 &87.82 &86.96  \\
\hline
\end{tabular}
\caption{ \label{tab_info}
    Comparison of touch exploration on dataset ABC. Numbers represent a ratio (\%) between CD after 5 actions and initial CD (with zero grasp). 
    }
\end{center}
% \vspace{-0.4cm} 
\end{table}
%-------------------------------------------------------------------------

\subsection{Evaluation on Policy}
\label{sec:exp3}

\begin{figure*}[tp!]
\begin{center}

  % \begin{minipage}{0.91\linewidth}
  % \centerline{\includegraphics[width=0.999\linewidth]{fig/touch/26650.png}}
  % \end{minipage}

  \begin{minipage}{0.91\linewidth}
  \centerline{\includegraphics[width=0.999\linewidth]{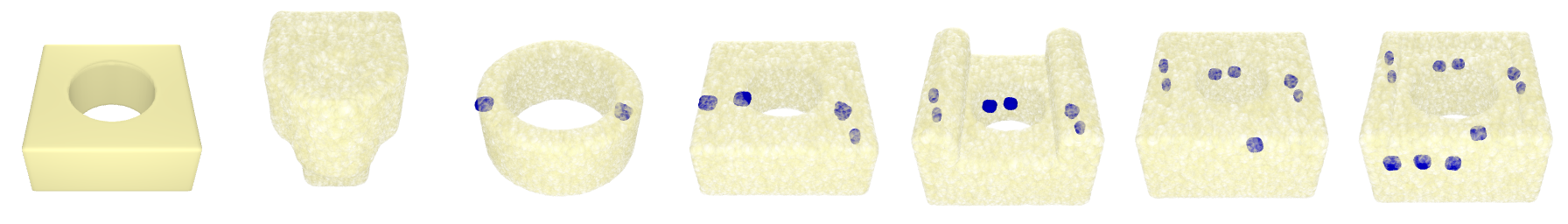}}
  \end{minipage}

  \begin{minipage}{0.91\linewidth}
  \centerline{\includegraphics[width=0.999\linewidth]{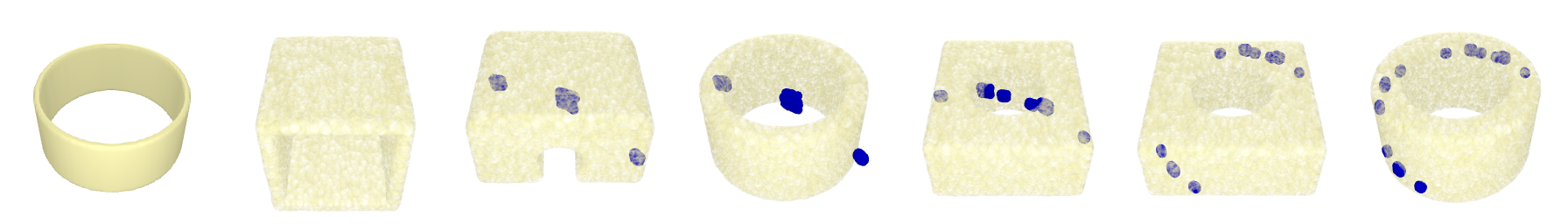}}
  \end{minipage}

  \begin{minipage}{0.91\linewidth}
  \centerline{\includegraphics[width=0.999\linewidth]{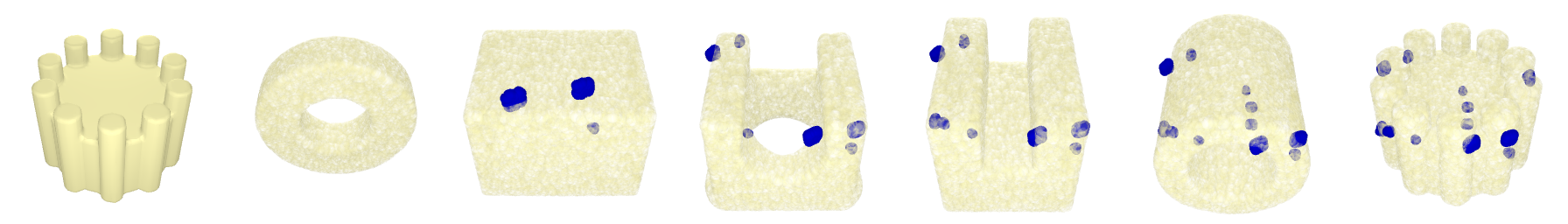}}
  \end{minipage}  

  \begin{minipage}{0.13\linewidth}
  \centering{\textbf{Ground truth}}
  \end{minipage}
  \begin{minipage}{0.13\linewidth}
  \centering{\textbf{\# Grasp: 0 }}
  \end{minipage}
  \begin{minipage}{0.13\linewidth}
  \centering{\textbf{\# Grasp: 1}}
  \end{minipage}
  \begin{minipage}{0.13\linewidth}
  \centering{\textbf{\# Grasp: 2}}
  \end{minipage}
  \begin{minipage}{0.13\linewidth}
  \centering{\textbf{\# Grasp: 3}}
  \end{minipage}
  \begin{minipage}{0.13\linewidth}
  \centering{\textbf{\# Grasp: 4}}
  \end{minipage}
  \begin{minipage}{0.13\linewidth}
  \centering{\textbf{\# Grasp: 5}}
  \end{minipage}
  
  % 9490，4136， 1480， 2271， 9680， 3320， 4437, 5437, 6901，11304
  \caption{ \label{fig:fig_num_touch}
           The evolution of the reconstructed shape with an increasing number of grasps (in the tactile only setting). Initially, limited information makes determining the overall global shape challenging, but with more grasp actions, our method effectively improves the reconstruction quality. The local points sampled on predicted touch charts (from captrued valid tactile images) are painted blue.
           }

\end{center}
% \vspace{-0.4cm} 
\end{figure*}

We evaluate our touch exploration policy model across both tactile only and visual-tactile settings. As \cite{activevt}, we set two baseline methods, Random and Even. The former policy selects one of the available actions at random while the latter results in uniform coverage of the target object. Furthermore, the Oracle policy is used to select the action which resulted in the best improvement, which is viewed as an upper-bound point of comparison as the true optimal policy cannot be computed in a reasonable time frame. As shown in Table \ref{tab_info}, the ratio between CD after 5 grasps and initial CD (with zero grasp) are reported. Our reconstruction method has a higher upper-bound point of reconstruction improvement, which validates the strong touch information fusion power of our model. The evolution of the reconstructed shape with an increasing number of grasps (in the grasp only setting) is illustrated in Figure \ref{fig:fig_num_touch}. We also visualize the touch points sampled on the touch charts (predicted from captured valid tactile images).  Initially, due to limited information, it is challenging to determine the overall global shape. As the number of grasps increases, our method gradually identifies both the global and local geometric structures.

%----------------------------------------------------------------------------------

%-------------------------------------------------------------------------

\subsection{Ablation Study}
\label{sec:exp4}

\begin{figure}[ht]
\begin{center}

  \begin{minipage}{0.98\linewidth}
  \centerline{\includegraphics[width=1\textwidth, height=0.2\linewidth]{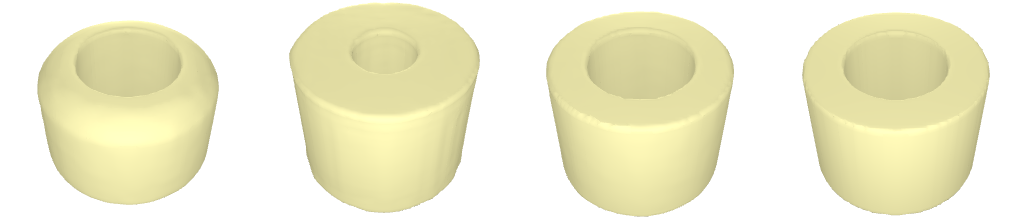}}
  \end{minipage}

  \begin{minipage}{0.98\linewidth}
  \centerline{\includegraphics[width=1\textwidth, height=0.2\linewidth]{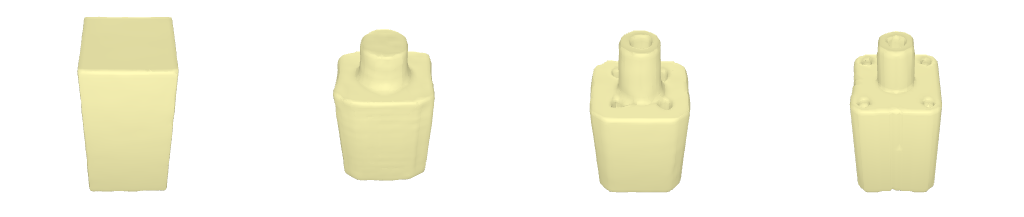}}
  \end{minipage}

  \begin{minipage}{0.24\linewidth}
  \centering{\textbf{T}}
  \end{minipage}
    \begin{minipage}{0.24\linewidth}
  \centering{\textbf{V}}
  \end{minipage}
  \begin{minipage}{0.24\linewidth}
  \centering{\textbf{V + T}}
  \end{minipage}  
\begin{minipage}{0.24\linewidth}
  \centering{\textbf{GT}}
  \end{minipage}
  
% \vspace{-0.4cm} 
  \caption{ \label{fig:fig_modal}
          visualization results among tactile only (T), visual only (V), and vision+touch (V+T) modalities.}
%  661, 2337, 5696, 5956, 1555, 4189
\end{center}
\vspace{-0.4cm} 
\end{figure}

We design the ablation study to further validate the necessity of our proposed reconstruction modules. The tactile images are captured through 5 random grasps. As shown in Table \ref{tab_ablation}, we add our proposed modules one by one to validate that each sub-module succeeds to improve the performance. We also train a vision-conditioned diffusion model (V represents the visual only setting) with a contrastive visual encoder. The evaluation results in different modes validate that our method can effectively integrate visual and tactile information to achieve a better reconstruction performance. We also visualize the 
results among tactile only, visual only, and visual-tactile modalities in Figure \ref{fig:fig_modal} (corresponding to rows 3, 5, and 7 in Table \ref{tab_ablation})

% %-----------------------------------------------------------------------
% \begin{table}[htbp]
% % \vspace{-2.0em}\\\
% \renewcommand{\arraystretch}{1.0}
% \begin{center}
% \begin{tabular}{| c | c | c | c | c | c |}
% \hline
% \textbf{\makebox[0.067\textwidth][c]{Mode}} &\textbf{\makebox[0.07\textwidth][c]{Touch}} &\textbf{\makebox[0.067\textwidth][c]{CL}}  &\textbf{\makebox[0.067\textwidth][c]{Fusion}} &\textbf{\makebox[0.07\textwidth][c]{CD $\downarrow$}} \\
% \hline  
% \multirow{4}{*}{\makecell{\textbf{{T}}}} &Points  &\XSolidBrush &\XSolidBrush &4.925    \\
%  &Charts  &\XSolidBrush &\XSolidBrush &4.430    \\
%  &Charts  &\Checkmark &\XSolidBrush &3.298    \\
%  &Charts  &\Checkmark &\Checkmark &\textbf{3.134}    \\
%  \hline
%  \multirow{2}{*}{\makecell{\textbf{{V}}}} &\XSolidBrush  &\XSolidBrush &\XSolidBrush &2.242  \\
%   &\XSolidBrush  &\Checkmark &\XSolidBrush &2.068  \\
% \hline
% \textbf{T+V} &Charts  &\Checkmark &\Checkmark &\textbf{1.304}  \\
%  \hline
% \end{tabular}
% \caption{ \label{tab_ablation}
%     ablation study results on dataset ABC. CL represents the contrastive encoder and Fusion represents the touch shape fusion module respectively.
%     }
% \end{center}
% \vspace{-0.4cm} 
% \end{table}
% %-------------------------------------------------------------------------

%-----------------------------------------------------------------------
\begin{table}[htbp]
% \vspace{-2.0em}\\\
\renewcommand{\arraystretch}{1.0}
\begin{center}
\begin{tabular}{| c |  c | c | c | c |}
\hline
\textbf{\makebox[0.08\textwidth][c]{Mode}}  &\textbf{\makebox[0.08\textwidth][c]{CL}}  &\textbf{\makebox[0.08\textwidth][c]{Fusion}} &\textbf{\makebox[0.08\textwidth][c]{CD $\downarrow$}} \\
\hline  
\multirow{3}{*}{\makecell{\textbf{{T}}}} &\XSolidBrush &\XSolidBrush &4.430    \\
  &\Checkmark &\XSolidBrush &3.298    \\
  &\Checkmark &\Checkmark &\textbf{3.134}    \\
 \hline
 \multirow{2}{*}{\makecell{\textbf{{V}}}}   &\XSolidBrush &\XSolidBrush &2.242  \\
    &\Checkmark &\XSolidBrush &2.068  \\
\hline
\textbf{T + V}   &\Checkmark &\Checkmark &\textbf{1.304}  \\
 \hline
\end{tabular}
\caption{ \label{tab_ablation}
    Ablation study results on dataset ABC. CL represents the contrastive encoder and Fusion represents the touch shape fusion module respectively.
    }
\end{center}
\vspace{-0.4cm} 
\end{table}
%-------------------------------------------------------------------------

% \subsection{Real World}

% \subsection{Limitations}

% pass.

%-------------------------------------------------------------------------

\section{Conclusion}
\label{sec:conclusion}

In this work, we present Touch2Shape, which leverages a touch-conditioned diffusion model to explore the target object and reconstruct the 3D shape through touch interaction. The generated latent vector from the diffusion model serves as a compact shape representation, guiding the policy model to predict beneficial touch locations for reconstruction based on potential missing areas. For shape reconstruction, we created a touch embedding module to condition the diffusion model and propose a touch shape fusion module to enhance the reconstructed shape. Extensive experiments demonstrate the effectiveness of our method, showcasing notable enhancements in reconstruction quality and shape optimization through grasping exploration compared to the state-of-the-art methods. 

There are also some future directions. Firstly, it's interesting to transfer Touch2Shape from the simulation to a real robot platform for object exploration and reconstruction. Secondly, the extension of Touch2Shape to full scene reconstruction through tactile exploration presents another fascinating dimension. Lastly, the study of integrating techniques like neural rendering could offer an intriguing pathway for leveraging active touch sensing to synthesis multi-view visual image.

%----------------------------------------------------------------------------------
% \vspace{-0.3cm} 

%----------------------------------------------------------------------------------

\section*{Acknowledgments}
% \textbf{Acknowledgments.}
This work is supported in part by the National Key Research and Development Program of China (No. 2022ZD0210500), the National Natural Science Foundation of China under Grant 62441216/62332019,  the Distinguished Young Scholars Funding of Dalian (No. 2022RJ01), and the Ningbo Major Research and Development Plan Project of China (No. 2023Z225).

{
    \small
    \bibliographystyle{ieeenat_fullname}
    \bibliography{main}
}
% WARNING: do not forget to delete the supplementary pages from your submission 
% \input{sec/X_suppl}

\end{document}